\documentclass[runningheads,a4paper]{llncs}
\usepackage[T1]{fontenc}

\usepackage{algorithm}
\usepackage{algorithmic}
\usepackage{graphicx}		% to insert figures
\usepackage{hypernat}		% PDF hyperreferences??
\usepackage{amsmath}
\usepackage{amssymb}		% to get all AMS symbols
\usepackage{comment}

\usepackage{booktabs}
\usepackage{multirow}
\usepackage{graphicx}
\usepackage{wrapfig}
\usepackage{subcaption} % For subfigure environment
\usepackage{etoolbox}
\usepackage{fontawesome5}

\makeatletter
\patchcmd{\@makecaption}
  {\scshape}
  {}
  {}
  {}
\makeatother

\usepackage[normalem]{ulem}
\usepackage{color}
\usepackage{soul}

\definecolor{greencheck}{RGB}{0,128,0} % Green color
\definecolor{redtimes}{RGB}{255,0,0}   % Red color

\usepackage{array}
\newcolumntype{P}[1]{>{\centering\arraybackslash}p{#1}}
\newcolumntype{M}[1]{>{\centering\arraybackslash}m{#1}}
\usepackage[font=small,labelfont=bf]{caption}
\usepackage{subcaption}
\usepackage{multirow}
\usepackage{float}     % For custom float types
\usepackage{wrapfig}
\usepackage{tabularx}
\usepackage{adjustbox} % For adjusting font size

\begin{document}
\title{Restorebot: Towards an Autonomous Robotics Platform for Degraded Rangeland Restoration\thanks{This work is supported by the National Science Foundation Robotics Initiative and the National Institute of Food and Agriculture project COLW-2020-08988 ``Autonomous Restoration and Revegetation of Degraded Ecosystems.}}
\titlerunning{Restorebot}
\author{Kristen Such \and
Harel Biggie \and 
Christoffer Heckman}
\authorrunning{K.Such et al.}
\institute{University of Colorado Boulder, Boulder CO 80309, USA \\
\email{christoffer.heckman@colorado.edu}\\}
\maketitle              % typeset the header of the contribution
\begin{abstract}
Degraded rangelands undergo continual shifts in the appearance and distribution of plant life. The nature of these changes however is subtle: between seasons seedlings sprout up and some flourish while others perish, meanwhile, over multiple seasons they experience fluctuating precipitation volumes and can be grazed by livestock. The nature of these conditioning variables makes it difficult for ecologists to quantify the efficacy of intervention techniques under study. To support these observation and intervention tasks, we develop RestoreBot: a mobile robotic platform designed for gathering data in degraded rangelands for the purpose of data collection and intervention in order to support revegetation. Over the course of multiple deployments, we outline the opportunities and challenges of autonomous data collection for revegetation and the importance of further effort in this area. Specifically, we identify that localization, mapping, data association, and terrain assessment remain open problems for deployment, but that recent advances in computer vision, sensing, and autonomy offer promising prospects for autonomous revegetation.

\end{abstract}

\section{Introduction}
Rangelands cover approximately 41\% \cite{nature_comms} of the Earth's land surface and support about 38\% of the global population. However, 20\% \cite{2006LivestocksOptions} of these rangelands are considered degraded due to poor land management, climate change, and increased development. Restoring these degraded areas is paramount for long-term sustainability and requires intervention which is intractable for humans to perform at scale due to the sheer volume of area. Autonomous robots present a promising alternative but despite advancements in autonomy and machine learning, there are still several challenges that need to be addressed if we are to develop a platform suitable for making meaningful interventions in these environments.

\begin{figure}[h]
    \centering
    \subfloat[\centering]{{\includegraphics[height=2.9cm]{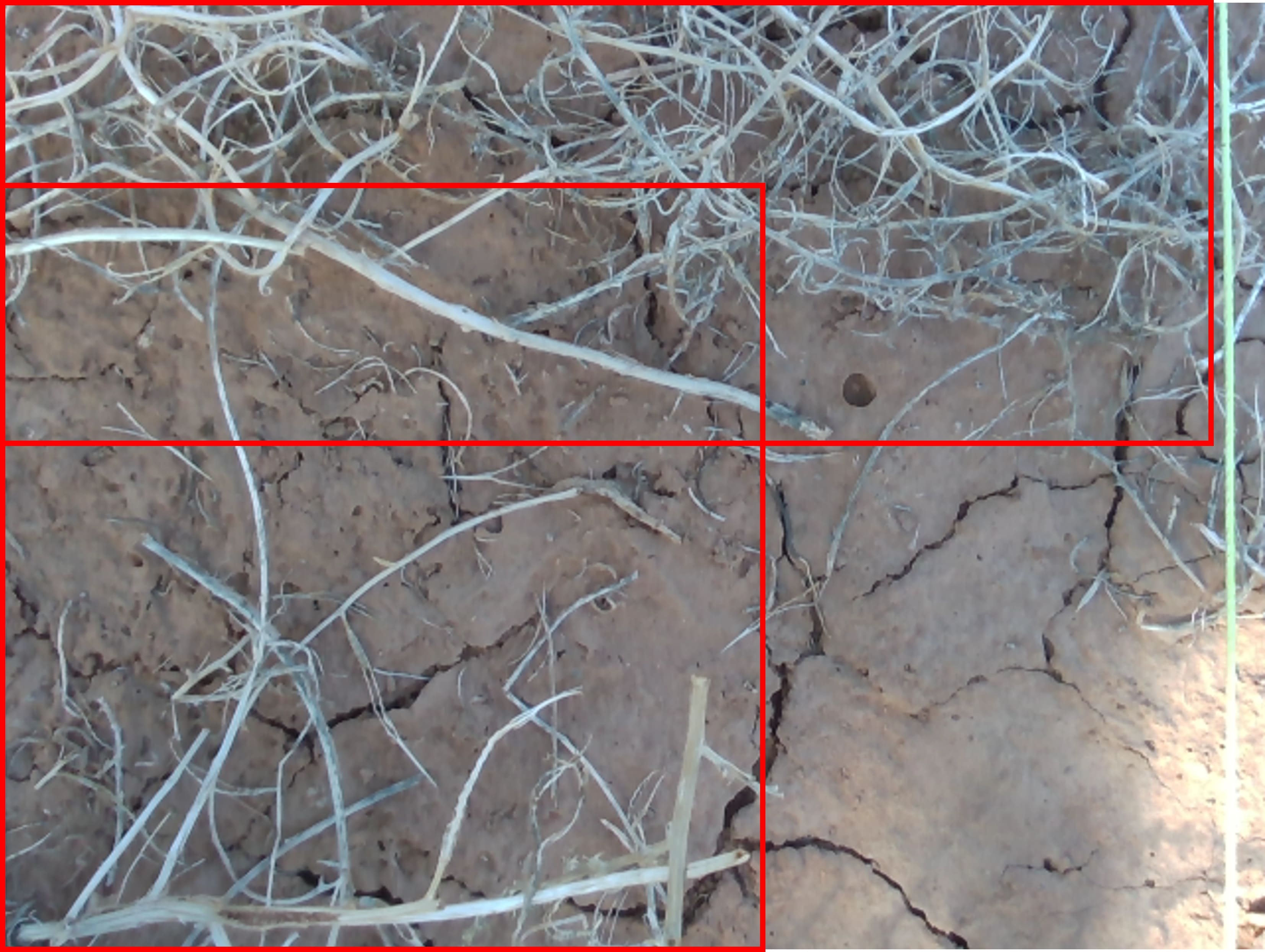} }}%
    \subfloat[\centering]{{\includegraphics[height=2.9cm]{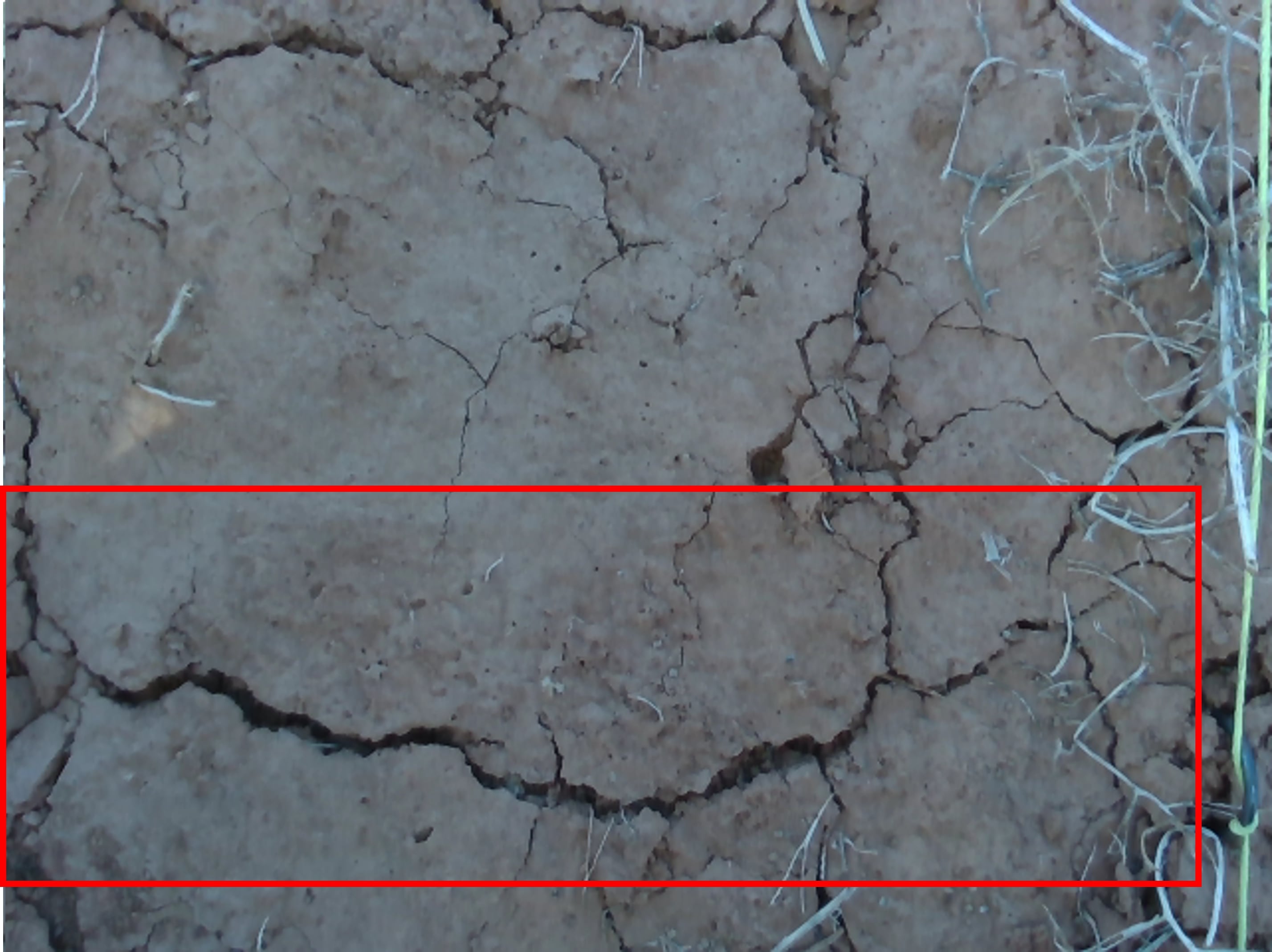} }}%
    \subfloat[\centering]{{\includegraphics[height=2.9cm]{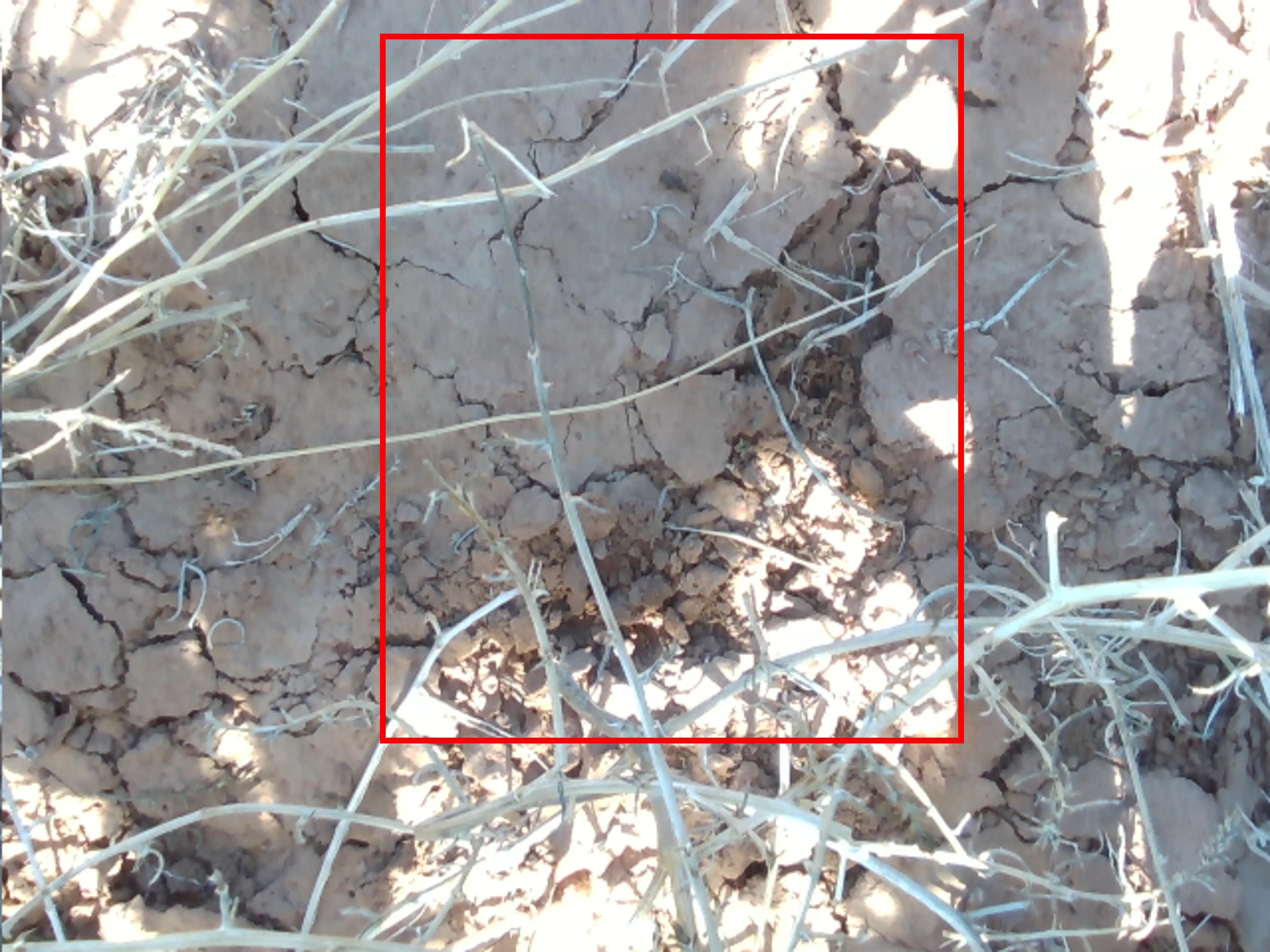}}}%
    \caption{Examples of bounding boxes created from dilations of hand-labeled microsite masks. These microsites were selected by our field ecology team as being potentially relevant to seedling establishment. (a) An example of a litter bounding box. (b) An example of a crack bounding box. (c) Example of a dip bounding box.}%
    \label{fig:microsite_examples}%
\end{figure}

Among restoration ecologists, degraded dryland ecosystems are considered the most prone to failure \cite{restoration_failure1,restoration_failure2} due to interactions between climate conditions (namely, warm temperatures and low precipitation) and alteration to ecosystem properties and processes resulting from land degradation as a consequence of improper land management. The exact reasons for decades of failed revegetation efforts in highly degraded rangelands are not well-understood by arid-land ecologists, although it is hypothesized that the failure of restoration projects in these environments has to do with microsite limitation \cite{microsite_limitation}. A microsite refers to the subtle features in the immediate vicinity (5--10cm radius) of a seedling or potential planting site. Microsite limitation occurs when there is a lack of sites with suitable soil conditions for seed germination. In degraded rangelands, water availability is the limiting resource, so sites that provide shading or favorable topography for water capture are likely to be the most important for seed establishment (see Fig. \ref{fig:microsite_examples}). The degradation in these environments worsens microsite limitation as the soil becomes bare and lacks the necessary plant shelter to protect seedlings from unforgiving environmental conditions. \\
\\
In this work, we detail our efforts in developing the foundations for an autonomous platform designed to restore degraded rangelands. We specifically detail opportunities and challenges for autonomous data collection in these environments while providing details on the ongoing challenges of plant localization, mapping, and data association across four field deployments in the Utah desert. In order to perform meaningful intervention it's critical to identify microsites across different seasons while tracking plant growth. This data can lead to a better understanding of where seedlings are more likely to grow and therefore where intervention is likely to succeed. However, performing centimeter localization while tracking plants in various growth states is problematic for existing techniques. We focus our work on techniques for performing microsite identification and tracking plants across different seasons. 
\section{Related Work}
% 1. Navigation in desert environments
For the purposes of uniquely identifying individual seedlings in this environment as they grow, we identify three key components: localization, image segmentation, and long-term data association. 

First, we require localization and mapping that is able to locate landmarks high accuracy and few observations. Connected to this is an extensive body of work on navigating in outdoor, unstructured environments \cite{Suger2015TraversabilityAF,stanley,urban_challenge}. Notably, these works are generally unconcerned with physically affecting the landmarks present in the field. In our experiments, we encountered similar complications as mentioned in \cite{agricultural_VI_SLAM,autonomous_agricultural_robotics}. We come to a similar conclusion to \cite{agricultural_survey_paper}, that even the state of the art when it comes to high-accuracy localization in agricultural robotics still does not meet the extremely tight margins of error that this work requires. We, therefore, aim to identify other grounding landmarks in the environment in order to have any hope of attempting this problem. 

% 2. Image segmentation, labelling, and feature assignment for visually redundant landmarks 
Second, we need to be able to identify vegetation and microsite conditions. In addition to our localization, we must then segment our images to identify individual plants. In particular we have relied heavily on Segment Anything \cite{segment_anything} which is the largest segmentation dataset to date and, as such incredibly well suited to zero-shot transfer to our more exotic application. 
 
Finally, we need to be able to correctly associate observations of the vegetation over very long time scales with very sparse observations. Given we cannot rely on localization alone to identify an individual plant, we require the data association to take the changing appearance of the plant into account. 

% 3. Semi-static SLAM
There are many existing methods that categorize objects as dynamic, static, or semi-static \cite{semanticFunction,dcnn}. These methods in their current state miss subtle appearance changes and result in misassociation of features to landmarks. Besides tracking these changes for the sake of that information, changes in an object's appearance may cause a break in the landmark's identification and subsequent localization. 

\begin{figure}[!htb]
    \centering
    \includegraphics[width=0.75\textwidth]{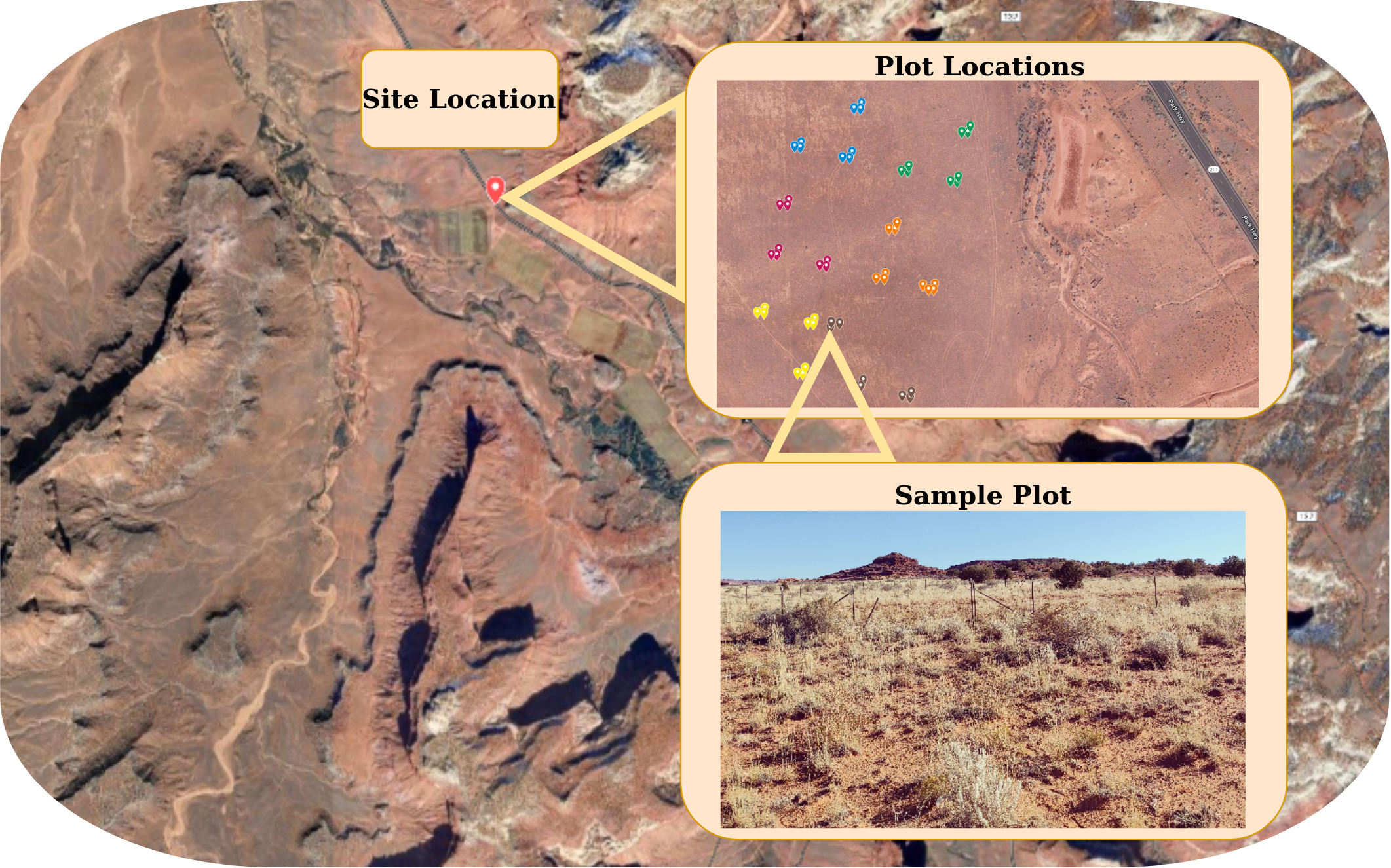}
    \caption{Satellite data \cite{googlemaps} of the Canyonlands research center and surrounding areas.} \label{fig:crc}
\end{figure}

An especially popular and elegant approach for establishing long-term feature-based maps is to instantiate multiple maps in parallel \cite{multimap,experience} based on repeated observations of a particular scene. In this approach, collections of landmarks from different sequences of motion through a scene (with corresponding pose constraints) are kept in memory with their varying feature-based appearances. On recall, the coexistence of these landmarks are collectively matched against the current collection of features for similarity, and those features which are successfully matched are used for estimating the robot position. While these techniques are successful in environments which have repetition in their appearance changes, they are inappropriate for sparse observations in which the map may change dramatically between observations.

\section{Canyonlands Research Center and Platform}
\subsection{Field Site}
The Canyonlands Research Center (CRC) is managed primarily by The Nature Conservancy (TNC)---a non-profit organization that developed out of the Ecological Society of America with the goal of preserving lands that harbor the diversity of plant and animal life. TNC worked with a private party \cite{carter2014assessing} to purchase a part of their private ranch and their herd for use in sustainable land management research. The site was desirable for TNC as it offered a window into the Arizona monsoon climate zone, had a history of existing climate data, several relict areas that provide baseline data, and the potential for ongoing cattle operation to inform policies for sustainable land usage. The CRC is a hub for several ongoing ecological research projects spanning many disciplines. 

\begin{wrapfigure}{l}{0.5\textwidth}
  \centering
  \includegraphics[width=\linewidth]{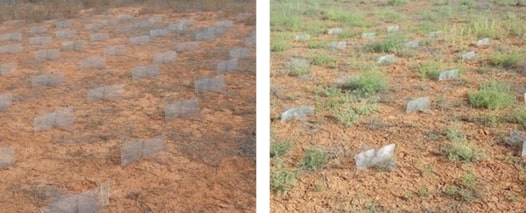}
  \caption{An example of ConMods in the field. These images were taken two years a part.}
  \label{fig:conmods}
\end{wrapfigure}

One technique developed by arid-land ecologists to create shelter and provide an anchor to slow soil erosion is known as a connectivity modifier (ConMod). ConMods are screens made of chicken wire fastened into crosses that are stapled into the ground (see Fig. \ref{fig:conmods}). They are about 30cm in diameter and end up forming small mounds around them in time as they help to catch soil that would have otherwise eroded. They are called as such because they help to increase the connectivity of vegetated patches. These ConMods are of particular interest to us, both as a potential avenue for future intervention and because they provide a static landmark to reference that remains unchanged despite seasonal changes. At CRC, multiple experimental plots are treated with these ConMods, contributing to the basis of our observational interest.

% With the exception of the first deployment in February 2021, we operated on one of 18 square plots measuring 10m$\times$10m in the CRC Restore Field. The field in which we permitted to work had received several treatments over the years. We surveyed 6 plots each of control, drill seeded, and ConMod-treated plots. 

\subsection{Our Platforms}
Over the course of deployments, we experienced a significant evolution in our platform type and payload. One of the first platforms we employed was focused purely on identifying gaps in sensing and compute payload, all packaged aboard a hand-dragged cart (``Restorecart''). This cart contained an Nvidia Jetson, two Intel RealSense D435 cameras (one forward-facing and one downward-facing), an Ouster OS1-64 lidar, and a LORD Microstrain 3DM-GX5-15 VRU.

\begin{figure}[!htb]%
    \centering
    \subfloat[\centering]{{\includegraphics[height=3.1cm]{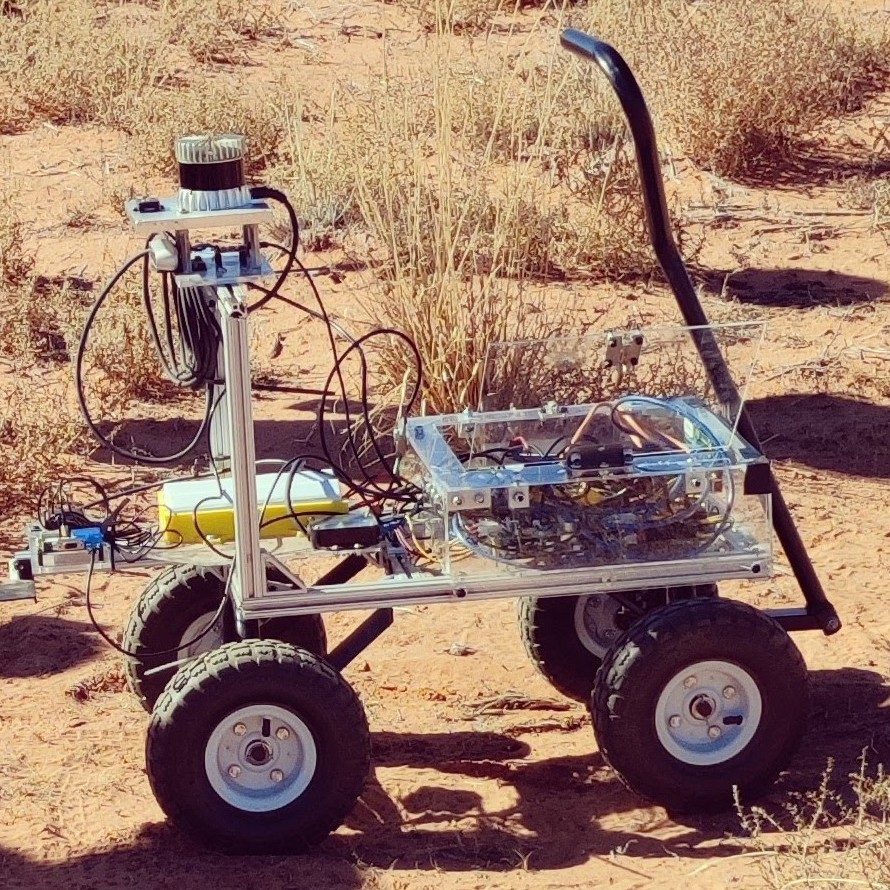} }}%
    \hfill
    \subfloat[\centering]{{\includegraphics[height=3.1cm]{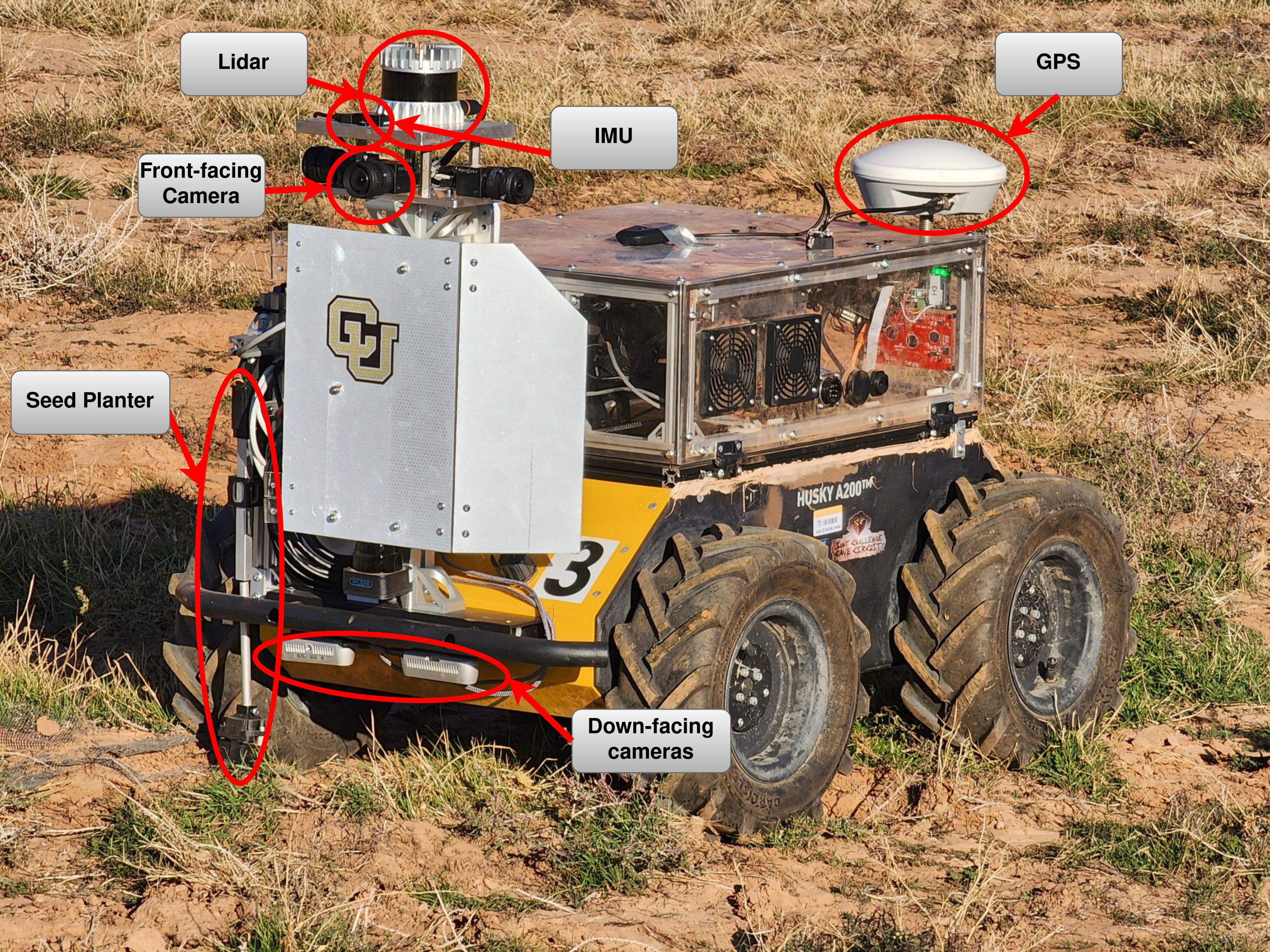} }}%
    \hfill 
    \subfloat[\centering]{{\includegraphics[height=3.1cm]{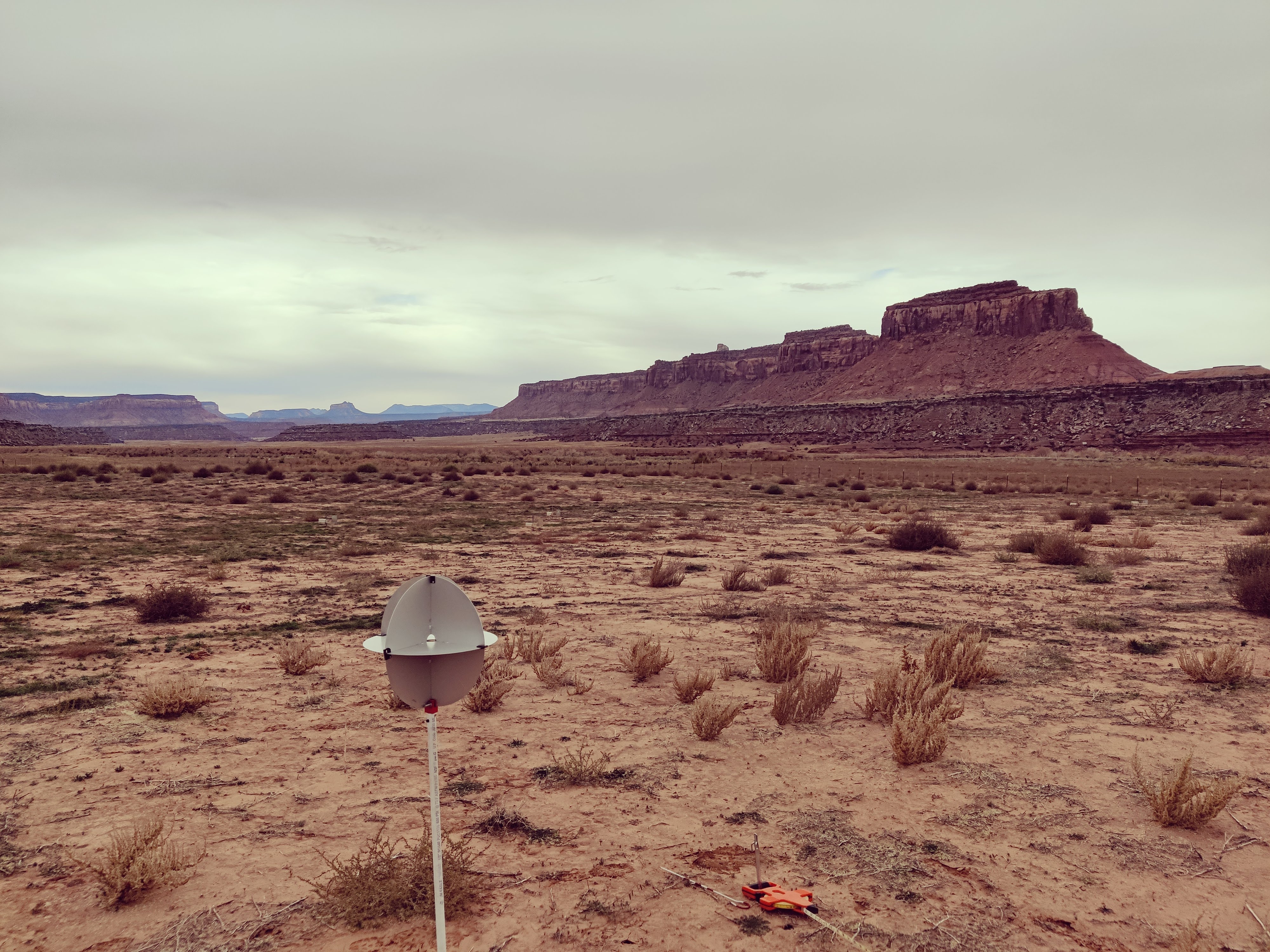} }}%
    \caption{(a) Restorecart (b) Restorebot with labeled sensors.(c) A radar reflector in the field in November 2021.}%
    \label{fig:platforms}%
\end{figure}

Over several deployments, the evolution of the platform led to an eventual embodiment on a mobile platform (``Restorebot'') shown in Figure \ref{fig:platforms}. On board this platform, a Clearpath Husky, are an Ouster OS1-64 lidar, a LORD Microstrain 3DM-GX5-15 VRU, a Trimble BX922 GPS with a Trimble AG25 antenna with 5cm RTK/RTX corrections, and a combination of Intel RealSense D435 and FLIR Blackfly cameras facing both forward and downward. The initial result of data collection and immediate post-processing yields GPS localization, lidar-intertial localization results (as well as the point clouds and inertial measurements themselves), a 15cm resolution Octomap \cite{hornung2013octomap} of the plots, point clouds, outward-facing RGB images and downward-facing RGB images. The on-board compute is an AMD Ryzen Threadripper which is capable of real-time object detection using a NVIDIA GTX 1650.

% \begin{wrapfigure}{l}{0.5\textwidth}
%   \centering
%   \includegraphics[width=\linewidth]{Content/Images/restorebot_labelled.jpg}
%   \caption{This is a robot.}
%   \label{fig:real-life-ex}
% \end{wrapfigure}
\section{Experiments}

\noindent \textbf{Field Experiments: Data Collection}

\begin{table}[h]
    \centering
    % \begin{tabular}{|c|c|*{5}{>{\centering\arraybackslash}m{1.2cm}|}}
    %     \hline
    %     Deployment Date & Platform & Point Clouds & IMU & Front-Facing Images & Down-Facing Images & RTK-GPS \\
    %     \hline

    \begin{tabular}{|c|c|*{5}{>{\centering\arraybackslash}m{1.2cm}|}}
        \hline
        Deployment Date & Platform & Point Clouds & IMU & Front-Facing Images & Down-Facing Images & RTK-GPS \\
        \hline
        February 2021 & Restorecart & \textcolor{redtimes}{\faTimes} & \textcolor{greencheck}{\faCheck} & \textcolor{greencheck}{\faCheck} & \textcolor{greencheck}{\faCheck} & \textcolor{redtimes}{\faTimes} \\
        \hline
        November 2021 & Restorecart & \textcolor{greencheck}{\faCheck} & \textcolor{greencheck}{\faCheck} & \textcolor{greencheck}{\faCheck} & \textcolor{greencheck}{\faCheck} & \textcolor{redtimes}{\faTimes} \\
        \hline
        May 2022 & Restorebot & \textcolor{greencheck}{\faCheck} & \textcolor{greencheck}{\faCheck} & \textcolor{greencheck}{\faCheck} & \textcolor{greencheck}{\faCheck} & \textcolor{greencheck}{\faCheck}  \\
        \hline
        November 2022 & Restorebot & \textcolor{greencheck}{\faCheck} & \textcolor{greencheck}{\faCheck} & \textcolor{greencheck}{\faCheck} & \textcolor{greencheck}{\faCheck} & \textcolor{greencheck}{\faCheck} \\
        \hline
    \end{tabular}
    \caption{Data products of the Restorecart and Restorebot Deployments.}
    \label{tab:my_label}
    \vspace{-5pt}
\end{table}

% Test
% \begin{table}[h]
%     \centering
%     \begin{tabularx}{\linewidth}{|>{\centering\arraybackslash}p{2.5cm}|>{\centering\arraybackslash}p{2.5cm}|*{5}{>{\centering\arraybackslash}X|}}
%         \hline
%         \multirow{2}{*}{Deployment \\ Date} & \multirow{2}{*}{Platform} & Point Clouds & IMU & Front-Facing Images & Down-Facing Images & RTK-GPS \\
%         & & & & & & \\
%         \hline
%         February 2021 & \adjustbox{max width=2.5cm}{Restorecart} & \emoji{cross-mark} & \emoji{check-mark} & \emoji{check-mark} & \emoji{check-mark} & \emoji{cross-mark} \\
%         \hline
%         November 2021 & \adjustbox{max width=2.5cm}{Restorecart} & \emoji{check-mark} & \emoji{check-mark} & \emoji{check-mark} & \emoji{check-mark} & \emoji{cross-mark} \\
%         \hline
%         May 2022 & \adjustbox{max width=2.5cm}{Restorebot} & \emoji{check-mark} & \emoji{check-mark} & \emoji{check-mark} & \emoji{check-mark} & \emoji{check-mark}  \\
%         \hline
%         November 2022 & \adjustbox{max width=2.5cm}{Restorebot} & \emoji{check-mark} & \emoji{check-mark} & \emoji{check-mark} & \emoji{check-mark} & \emoji{check-mark} \\
%         \hline
%     \end{tabularx}
%     \caption{Data products of the Restorecart and Restorebot Deployments.}
%     \label{tab:my_label}
% \end{table}
% located within the Bears Ears National Monument,

We deployed our ground platforms to a former cattle pasture at CRC outside Monticello, UT, over the course of two years. As the CRC is located on an active cattle ranch, it is a prime example of degraded rangeland.  

\noindent \textbf{February 2021 Deployment.} 
Our first deployment in February 2021 had the primary intention of hand-seeding Indian Rice Grass into microsites as we thought we could formulate the problem of robotic revegetation as a reinforcement learning problem. That is, we wanted to take as many actions as possible to explore the reward space of the revegetation problem.   

The experimental procedure involved an ecology student standing in the field alongside the cart inputting microsite locations into a tablet. We subdivided the plot into rows that were exactly the width of the cart's wheelbase and set quotas for each microsite type indicated to us by the ecologist as relevant; cracks, dips, and litter. At this point we were relying on visual-inertial odometry for our SLAM solution, namely, we employed \cite{orbslam} as we suffered a catastrophic hardware failure with the lidar upon arrival at the field site. 

However, we had to abandon the reinforcement learning formulation as no seedlings sprouted leading to a lack of reward signal. Later we found out that four consecutive years of ecological interventions in that particular field had also experienced zero evidence of plant recruitment. This suggests an unseen state that is critical to plant success in such highly degraded environments. Although our field ecologist counterparts cannot definitively explain the root cause of this failure, they point to a lack of microbial activity in the soil as a contributing factor. Simply put, they suspect that the soil in that plot was essentially inert.

\noindent \textbf{November 2021 Deployment.} 
That following fall, we received reports that seedlings had naturally established in a neighboring plot a few miles away. We deployed RestoreCart hoping that comparing the microsite conditions of that plot to our previous site might reveal quantitative information about revegetation in these environments. However, we learned that visual-inertial odometry placed into a reference frame using physical field markers was nowhere near accurate enough to help distinguish between individual seedlings. This is when we began trying to use the GPS reference frame. We did not have centimeter resolution GPS available that would be suitable for our cart, but we did have access to a Trimble Zephyr, a survey-grade GPS, which could collect extremely accurate GPS coordinates, albeit with the condition that we had to leave the tripod stationary for about two minutes for each GPS coordinate we wanted to take. As such, we had to be selective about what coordinates were most critical for revisiting the plot in the future.

The ecologists we worked with had already staked out several plots within this less degraded field site for us to investigate - there were 18 in total, 6 of each of the three treatments they were testing; drill seeded, ConMod, and control. Each plot was 10m x 10m as measured in the field. We installed rebar stakes in three out of four corners of each of these plots and used the Trimble Zephyr to get a highly accurate measurement of each of these corners. Then, as we had understood the extreme scarcity of lidar features available to us from the previous deployment, we installed radar reflectors on top of each rebar stake to help improve the localization result. That way, we could localize in reference to the highly accurate GPS coordinates. This ended up not working as well as we had hoped, as the mounts we had manufactured for the purpose of mounting the radar reflectors to the rebar stakes almost immediately failed in the intense winds in the field. As such, the centers of the radar reflectors moved too much for them to provide the level of accuracy required in the GPS frame. 

After installing the rebar and surveying the plot coordinates, the experimental procedure consisted of slowly dragging the cart in tight rows, trying to ensure that the down-facing camera covered as much of the plot area as possible. We were careful not to step within the plot itself as this would disturb the microsites. 

\noindent \textbf{May 2022 Deployment.} 
For the next deployment, in May 2022, we received significant hardware upgrades. We not only switched to the Restorebot platform, but we also obtained a high-accuracy survey-grade GPS. We also added a seeding mechanism consisting of a spring-loaded hand seeder and a simple linear actuator on the bottom of the robot. Besides greatly reducing the amount of physical labor required in our fieldwork, the quality of our down-facing images greatly improved as the amount of motion blur due to the vibration of the platform was greatly dampened. We also realized that, although the only lidar features in the environment (in most instances) are on the ground, we understood from our previous deployment that this actually did not significantly deteriorate our localization estimates, only our ability to compare between seasons as we were unable to leverage even low-overlap iterative closest point \cite{stechschulte2019robust} or other lidar-based methods for refinement of the transformation between seasonal observations. As we had automated the planting process and it was the appropriate time of year to lay seed, we also seeded Indian Rice Grass in four of the six ConMod plots during this deployment.

\begin{figure}[htb]
    \centering
    \includegraphics[width=11.5cm]{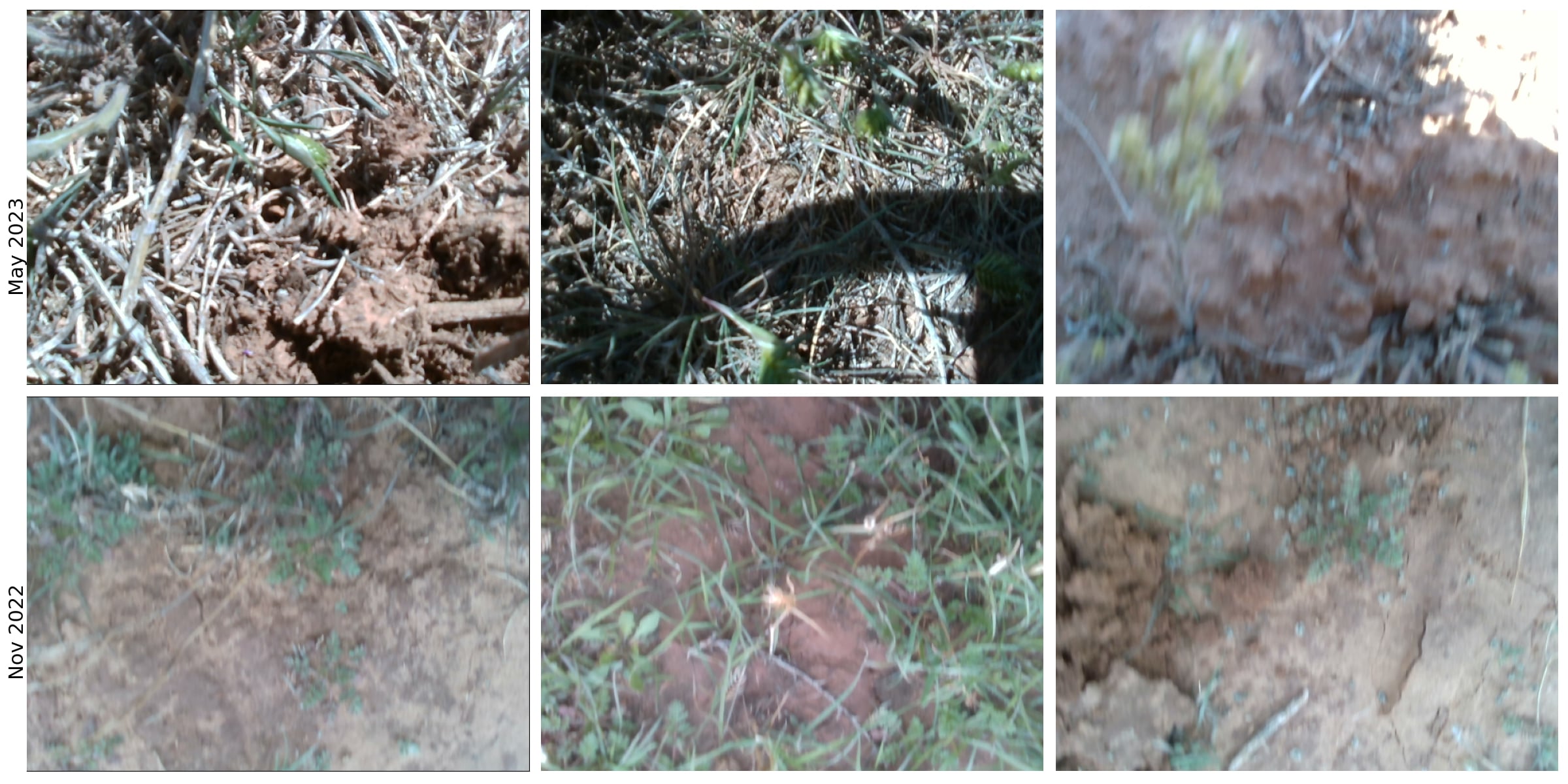}
    \caption{Comparison of frames between two seasons that were estimated to be less than 10cm apart. Clearly, the uncertainty in the localization is too large to rely on localization alone for unique plant identification. Also note the shadows of the robot, the differences in lighting, and the visual redundancy.} \label{fig:cross_seasonal_examples}
    %\vspace{-15pt}
\end{figure}

The experimental procedure during this deployment was similar to November 2021 except we teleoperated the robot in a similar pattern over each of the plots, and did not install the radar reflectors on the corners of the plot. Once we had completed our surveying, we returned to seed as many plots as we could, starting with the ConMod plots as we had reason to believe that they were the most likely to successfully recruit the seeds. In the original spirit of exploring the reward space, we seeded roughly in the center 1m $\times$ 1m grids that we staked out using string. 

\noindent \textbf{November 2022 Deployment.} 
In November 2022, our goal was to collect the data that we needed for cross-seasonal data association with a secondary objective of observing seedlings that might have been established from our efforts in the previous deployment. The only change in our platform was the addition of another down-facing camera to increase the high-resolution coverage of the plots. 

The experimental procedure was identical to that of May 2022, except we did not plant seeds due to the time of year. During this deployment, collected additional data after collecting the survey data of the plots that focused on characterizing problematic aspects of the terrain which had so far prevented us from running autonomous navigation experiments out of safety concerns for our platform. These hazards include abandoned field markers and fencing which are often undetectable with lidar and sometimes are even obscured in the camera's field of view. Such hazards were even difficult for our field team to avoid.

% \begin{figure}%
%     \centering
%     \subfloat[\centering]{{\includegraphics[height=2.5cm]{Content/Images/plantMask_ex.png} }}% 
%     \hfill
%     \subfloat[\centering]{{\includegraphics[height=2.5cm]{Content/Images/segment_anythin_ex.png}}}%
%     \vspace{-5pt}
%     \caption{(a) Our experimental platform. (b) An overview of our data processing approach \cite{liosam2020shan}\cite{segment_anything} (c) Example of a plant identification from downward facing images.}%
%     \label{fig:example}%
% \end{figure}

\noindent \textbf{Field Experiments: Data Collection.}
For each down-facing image we collected, we segment the image using \cite{segment_anything}. We then label each of the masks using a custom-trained binary classifier. Then, given a labeled set of pixels corresponding to a landmark of interest, we obtain a 3D coordinate of the landmark by projecting the centroid and bounding box of the mask using a linear camera model. We apply $k$-means clustering and sparse feature matching, which enabled us to associate the 3D coordinates with individual plants within one season. Together, this process yields a map of the vegetation, environmental features of interest, and the robot's trajectory using \cite{liosam2020shan}.

\begin{figure}[htb]%
    \centering
    \subfloat[\centering]{{\includegraphics[height=3.5cm]{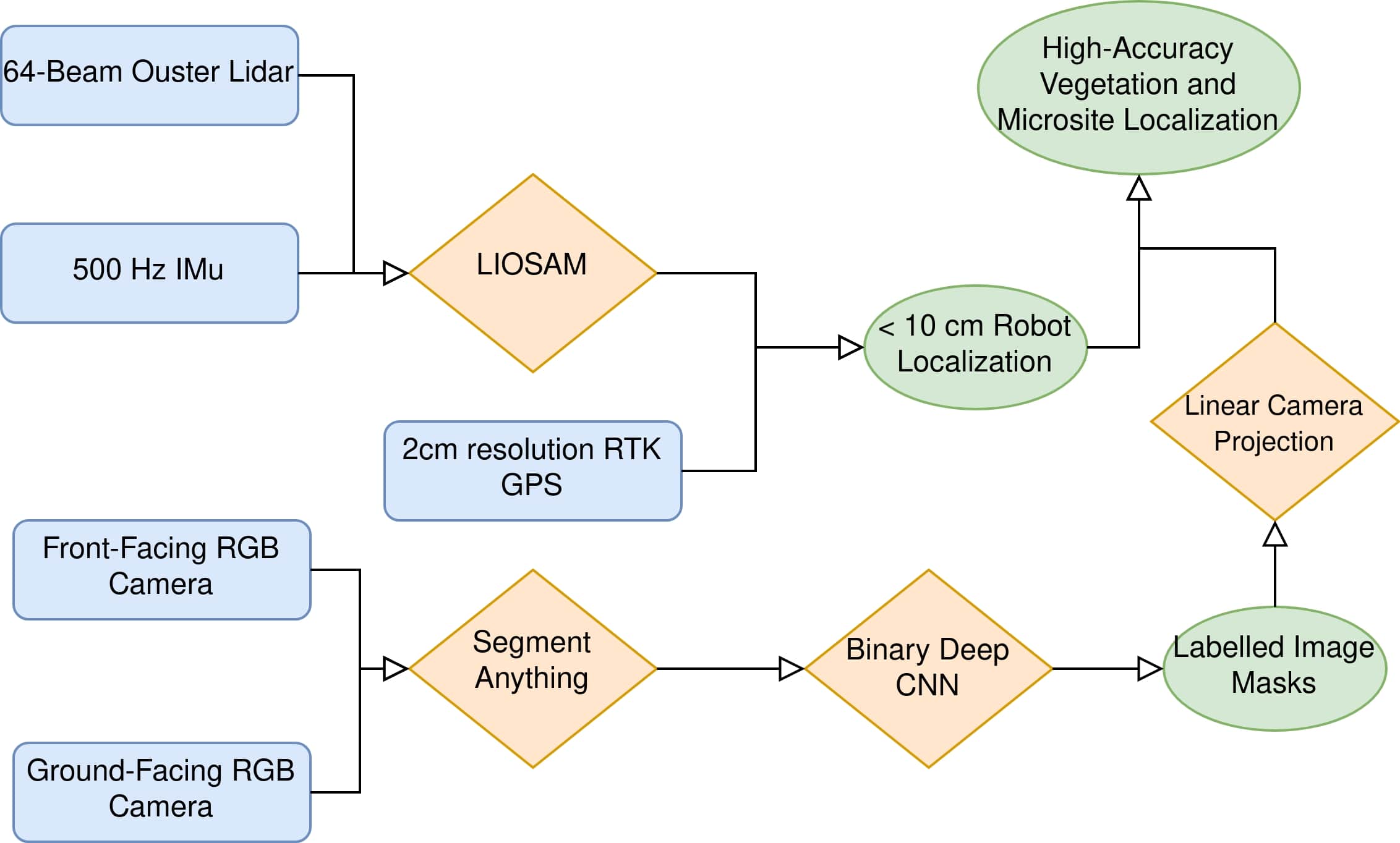} }}%
    \hfill
    \subfloat[\centering]{{\includegraphics[height=3.5cm]{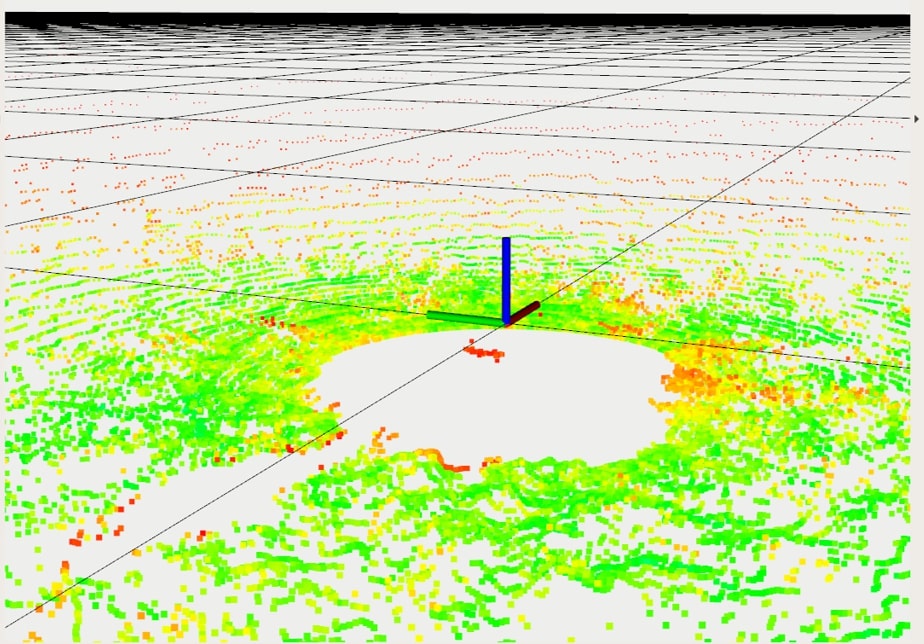} }}%
    \caption{(a) Data processing pipeline. (b) An example of a typical lidar scan in the field. Note the lack of distinguishing landmarks which would help to uniquely identify a plot.}%
    \label{fig:pipeline_and_lidar}%
    \vspace{-7pt}
\end{figure}

Similarly, we segment the front-facing camera frames to identify what constitutes the ground in the image. We again apply \cite{segment_anything} to this sub-mask to find ConMods and larger vegetation such as shrubs. For shrubs in particular, we find the ray box intersection of the 2D pixel bounding box by traversing the Octree nodes. We were able to convert all of these localized landmarks into the GPS frame using the Vincente formula and the high-accuracy GPS measurements, see Fig. \ref{fig:pipeline_and_lidar} (a). 

% \begin{wrapfigure}{l}{0.5\textwidth}
%   \centering
%   \includegraphics[width=\linewidth]{Content/Images/dataprocessing_pipeline.png}
%   \caption{Data Processing Pipeline}
%   \label{fig:real-life-ex}
% \end{wrapfigure}

For cross-seasonal data association, we iterate through each clustered object identified as vegetation, we find all vegetation objects that fit within our estimated localization covariance and take the closest match in terms of location and vegetation type. Given we find a cross-seasonal GPS match between seasons, we take all the frames within a 2m radius and find all the surrounding vegetation and microsites. 

However, this procedure on its own does not yield any convincing instances of cross-seasonal data association (see Fig. \ref{fig:cross_seasonal_examples}). We identify the reasons for this failure in terms of localization, image segmentation and labeling, and long-term data association as well as some proposed solutions. 

\noindent \textbf{Procedure: Localization and Mapping.} 
Our field site has the advantage of having access to GPS measurements as it consistently has open sky conditions, but is extremely feature sparse and exposed to direct sunlight most of the time, see Fig. \ref{fig:pipeline_and_lidar} (b). As reported in \cite{liosam2020shan}, the RMSE (root mean squared error) translation error compared to GPS can be expected to be off by as much as 1 meter. Although this is exceptional in most contexts, this leaves much room for misassociation of individual plants which are more or less visually indistinguishable even to a human. Even state-of-the-art visual-inertial localization and mapping algorithms suffer a similar problem due to the extreme visual redundancy and sparseness of the environment. Indeed, if we are to rely strictly on localization for cross-seasonal data association of plants, the error in the localization that would be required is so far unachievable, at least at our field site. Furthermore, given that there are dramatic changes that occur across seasons in the appearance of the vegetation and that we only have very sparse observations of the field site, it would be very wrong to assume you could overcome such shortcomings in localization by merely using simple feature comparison to uniquely identify individual plants. 
% \begin{figure}%
%     \centering
%     \subfloat[\centering]{{\includegraphics[height=3.5cm]{Content/Images/Picture1.png} }}%
%     \hfill
%     \subfloat[\centering]{{\includegraphics[height=3.5cm]{Content/Images/dataprocessing_pipeline.png} }}%
%     \caption{(a) An example of a typical lidar scan in the field. Note the lack of distinguishing landmarks which would help to uniquely identify a plot. (b) An example of a plant image and its masked counterpart segmented using \cite{segment_anything}.}%
%     \label{fig:lidarscsan_segmentanything}%
%     %\vspace{-22pt}
% \end{figure}
Thus far, our proposed solution to this problem is to semantically identify static landmarks in the environment and associate our semi-static landmarks in relationship to the static landmarks. In our case, the best candidate for the static landmarks is the ConMods. Although their surroundings change over time as they accumulate soil and litter, the panels themselves remain in the same place unless or until they are removed.
\begin{figure}%
    \centering
    \subfloat[\centering]{{\includegraphics[height=2.6cm]{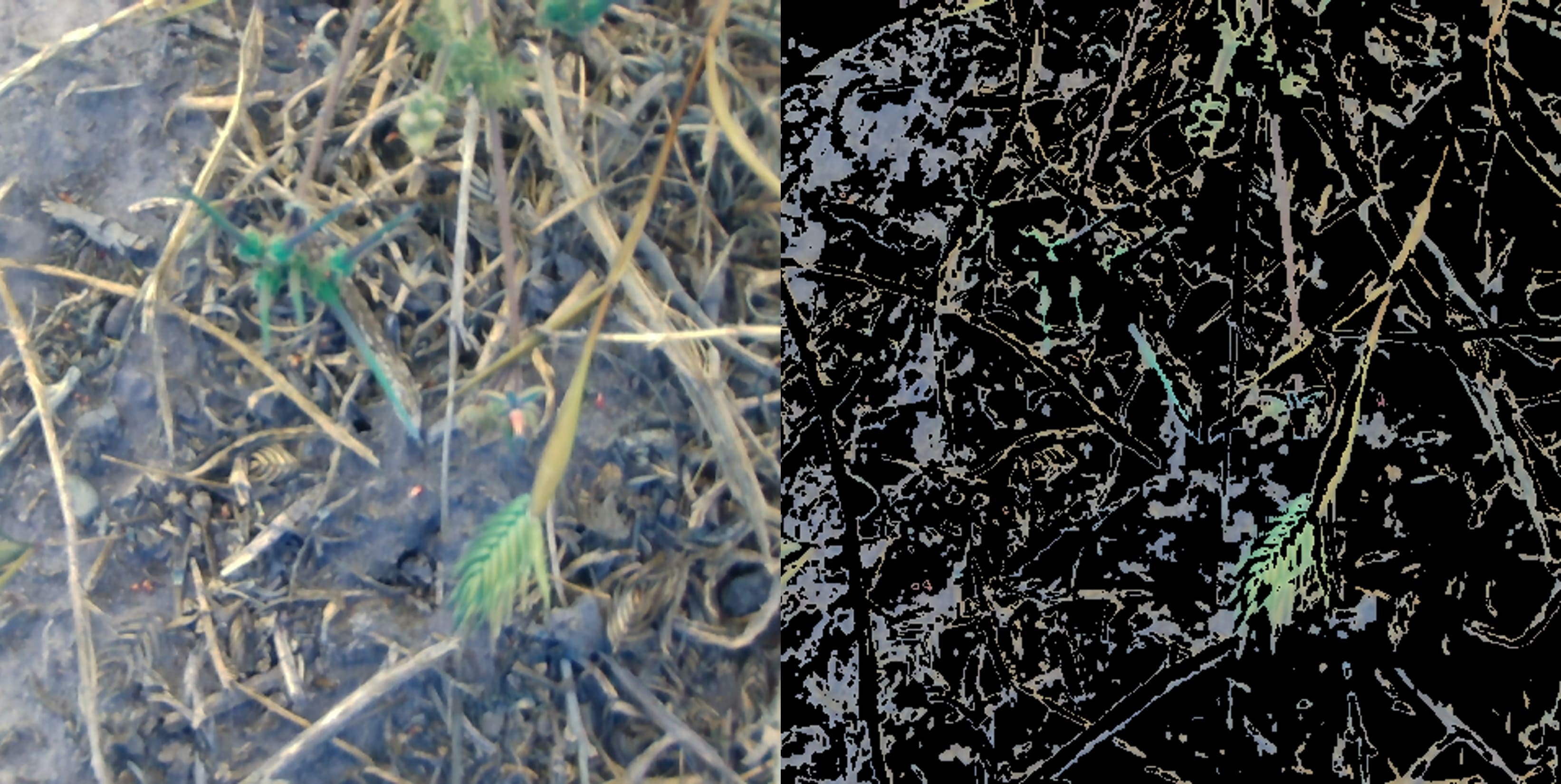} }}%
    \hfill
    \subfloat[\centering]{{\includegraphics[height=2.6cm]{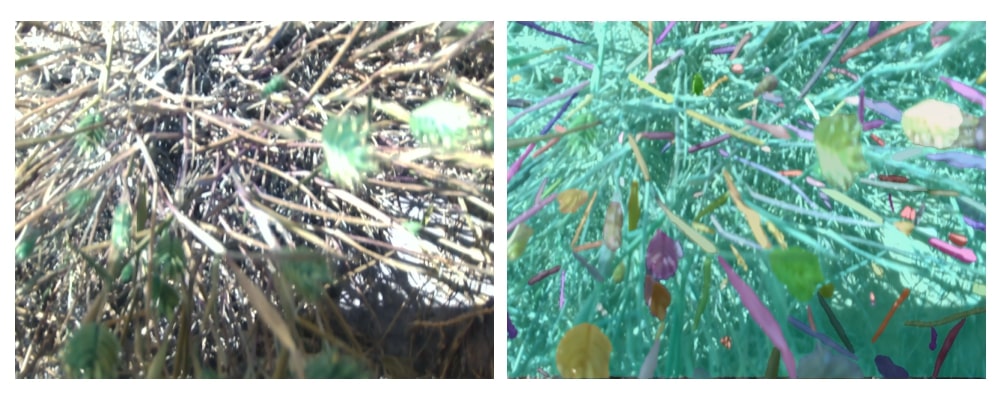} }}%
    \caption{(a) Here is an example of an image alongside its plant mask. We used  \cite{segment_anything} and then applied a color filter. As you can see, there is still a need for further filtering to accurately identify plant centroids. (b) An example of a segmented image using \cite{segment_anything}.}%
    \label{fig:segmentanything_plantMask}% 
    \vspace{-8pt}
\end{figure}

\begin{figure}
    \centering
    \includegraphics[width=\textwidth]{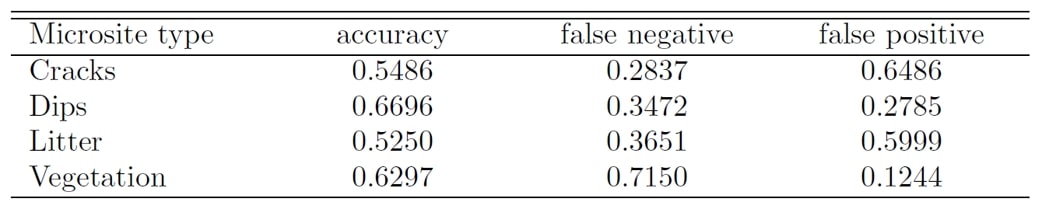}
    \caption{Our microsite and vegetation detection results on our validation set. We hand-labeled masks of each microsite type and the vegetation. We had 363 examples of litter, 441 examples of cracks, 407 examples of dips, and 174 examples of vegetation. We trained \cite{detectron2} over 40,000 iterations.} \label{fig:detectron_results}
    \vspace{-15pt}
\end{figure}

\noindent \textbf{Procedure: Image Segmentation and Labeling.}
The recent developments in image segmentation have greatly improved our results even within the relatively short course of this project. However, we see that further processing is required on these masks to be able to use a binary classifier for the purposes of identifying microsites or vegetation in our environment (see Fig. \ref{fig:segmentanything_plantMask} (b)). We could successfully identify objects of interest using Mask RCNN \cite{detectron2} on manually labeled images (see Fig. \ref{fig:detectron_results}), implying that the bulk of the work remaining to successfully identify microsites and vegetation lies in segmentation rather than labeling. 

%\noindent \textbf{Procedure: Cross-Seasonal Data Association} 

% \subsection{Planned Experiments}
% We are currently developing a pipeline to use BetterFaster on a toy problem in our lab in order to simplify a few of these challenges. Specifically, we set up an analogous experiment in our motion capture space in which we placed boxes to represent the ConMods and places patches of tape on the floor to represent the patches of vegetation. We used the motion capture space to get the ground-truth location of these landmarks and then drove the robot around the ``plot'' to observe the patches of tape. Then we would stop recording and localization, change the tape patches (the patches could increase or decrease in size and/or change color, but the centroid of the patch remained constant) according to a prescribed growth model, and repeat over multiple ``seasons''. That is, we defined a set of three such passes as a year, with each pass representing a season. The ``plants'' followed a Gaussian distribution of persistence and growth.

% On this analogous problem we have been further developing our data processing pipeline and proposed framework for cross-seasonal data association. \textcolor{red}{BetterFaster stuff here?}

% our experimental data in order to improve our cross-seasonal tracking accuracy. We believe that the particle filtering will improve the localization result of the landmarks which will increase the number of cross-seasonal matches we find and better characterize the uncertainty in these matches. 

\section{Future Work}
% %this should be one page 
% \noindent \textbf{Semi-static landmark localization.}
%  We find that by modeling the correlations between features within a clique, BetterFaster SLAM is more resistant to occlusions and missed detections. By examining individual cliques as opposed to updating entire maps, we achieve a higher mean accuracy and better landmark location estimate compared to ground truth. We find that the improved performance of BetterFaster can be explained by the fact that our method preserves more of the measurements---indeed, in our simulations, BetterFaster integrated about 33\% more measurements than MultiMap-based methods. This reinforces our hypothesis that discarding observations of dynamic objects ultimately reduces the accuracy of SLAM results in semi-static environments. %, with BetterFaster integrating 69\% of measurements and MultiMap integrating 52\%. 

\noindent \textbf{Cross-seasonal data association.}
We are especially motivated to implement a joint-persistence filter-based method for cross-seasonal data association using our multiple seasons of data collected from the Canyonlands Research Center. We consider expanding on the approach of \cite{bettertogether} and \cite{Nobre} by introducing some transition function that could account for the changes in the appearance of the landmarks without dropping the estimated posterior. That is, we believe that by integrating a knowledge of how features associated with plants could change over time, we could then anticipate these physical changes in appearance and maintain our estimate of the location of the landmark.

\noindent \textbf{Semi-Degraded Field Sites.}
We also found that revegetation is nontrivial in less degraded field sites. In May 2021, we deployed to a non-degraded pasture intending to observe the differences between conditions in a healthy field site versus a degraded one. Upon arrival, we observed catastrophic localization failures due to tall grasses blowing in the wind, leading to failed geometric consistency. The thick swaying grass was not identified as a dynamic object by our lidar-based localization system, and as such the robot appeared to be like a boat on a tide: moving in time with the wind. This suggests to us that if we want to expand the use of such a platform to other kinds of rangeland environments, we would have to integrate semantic terrain identification into our terrain traversability stack.

\section{Conclusion}
Reflecting on our deployments to the Canyonlands Research Center in Monticello, UT, we identify long-term data association with sparse observation as the key challenge facing autonomous revegetation. We focus on solutions to long-term data association that are resilient to semi-static changes, e.g., subtle differences in visual appearance over observations that are temporally distant. Towards this end we introduce a pipeline for field operations and mapping that relies on a joint probability filter that operates on features which change over time according to an a priori model.
\vspace{-15pt}

\subsubsection{Acknowledgements} 
We would like to acknowledge the many people who helped us in our Field Work. In particular, we thank our ecology team; Nichole Barger and Claire Karban, who gave us access to our field sites and offered invaluable insights into field ecology and aridland restoration. We also thank James Watson who assisted us in the last two of our deployments.
\vspace{-10pt}
\bibliographystyle{splncs04}
\bibliography{references}




% LNCS Homepage, \url{http://www.springer.com/lncs}. Last accessed 4
% \end{thebibliography}
\end{document}